\documentclass[sigconf]{acmart}
\usepackage{hyperref}
\AtBeginDocument{%
  }

\setcopyright{acmlicensed}
\copyrightyear{2025}
\acmYear{2025}
\acmDOI{XXXXXXX.XXXXXXX}
\acmConference[ACM SIGKDD]{2nd Workshop on
Agentic AI for Enterprise}{August 3–7, 2025}{Toronto, ON, Canada}
\acmISBN{978-1-4503-XXXX-X/2018/06}

\begin{document}

\title{Data-to-Dashboard: Multi-Agent LLM Framework for Insightful Visualization in Enterprise Analytics}

\author{Ran Zhang}
\affiliation{%
  \institution{Boston University}
  \city{Boston}
  \state{MA}
  \country{USA}}
\email{ran0925@bu.edu}

\author{Mohannad Elhamod}
\affiliation{%
  \institution{Boston University}
  \city{Boston}
  \state{MA}
  \country{USA}}
\email{elhamod@bu.edu}

\renewcommand{\shortauthors}{Zhang and Elhamod}

\begin{abstract}

The rapid advancement of LLMs has led to the creation of diverse agentic systems in data analysis, utilizing LLMs' capabilities to improve insight generation and visualization. In this paper, we present an agentic system that automates the data-to-dashboard pipeline through modular LLM agents(See Fig. \ref{fig:agenticmodule})  capable of domain detection, concept extraction, multi-perspective analysis generation, and iterative self-reflection. Unlike existing chart QA systems\cite{perez2025llm,wu2024automated,sahu2024insightbench,xu2023chartbench}, our framework simulates the analytical reasoning process of business analysts by retrieving domain-relevant knowledge and adapting to diverse datasets without relying on closed ontologies or question templates.
We evaluate our system on three datasets across different domains. Benchmarked against GPT-4o with a single-prompt baseline, our approach shows improved insightfulness, domain relevance, and analytical depth, as measured by tailored G-Eval metrics\cite{liu2023gevalnlgevaluationusing} and qualitative human assessment.

This work contributes a novel modular pipeline to bridge the path from raw data to visualization, and opens new opportunities for human-in-the-loop validation by domain experts in business analytics. All code can be found here: \url{https://github.com/77luvC/D2D_Data2Dashboard}

\end{abstract}




\maketitle

\section{Introduction}

\begin{figure}
    \centering
    \includegraphics[width=1\linewidth]{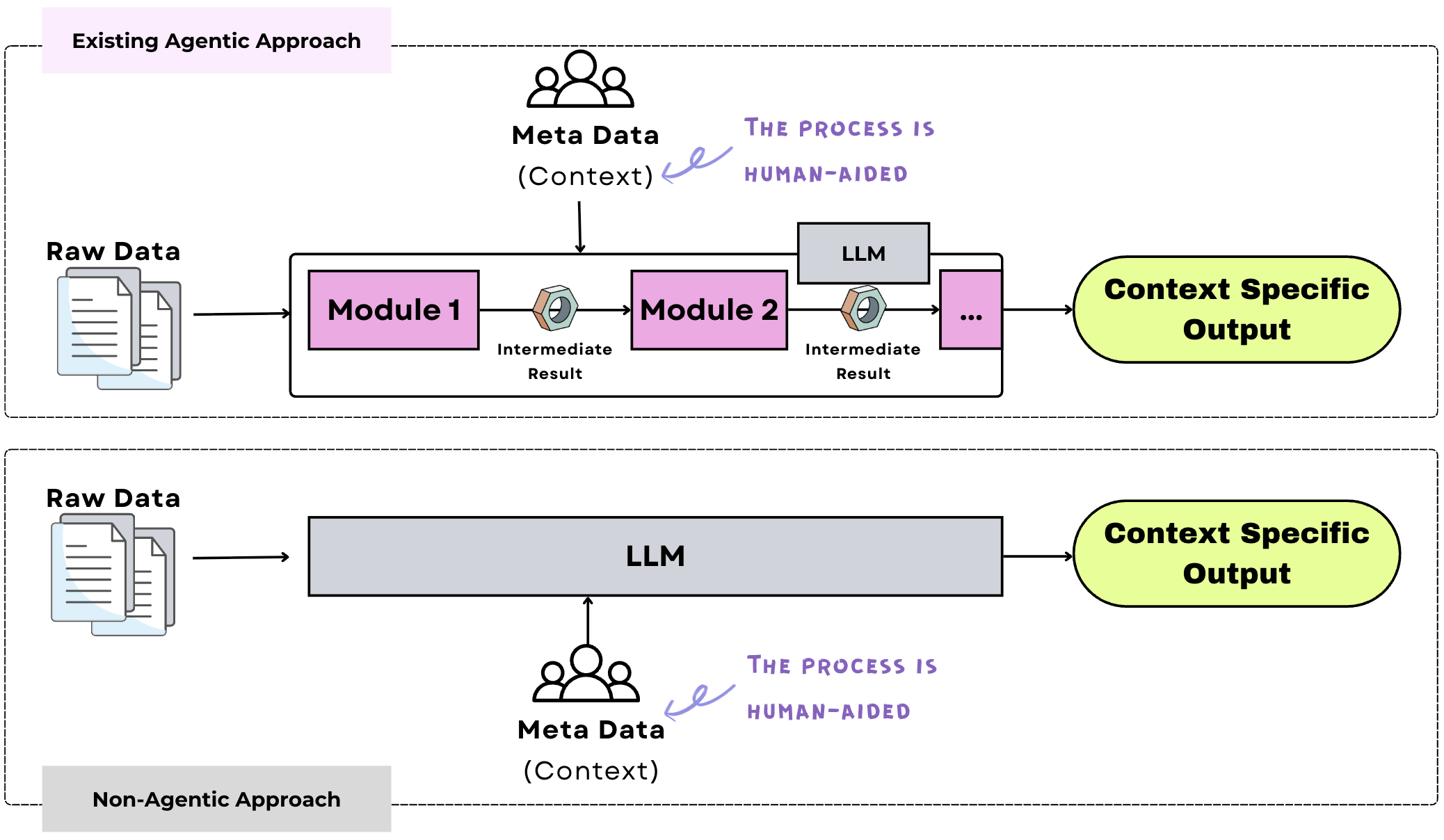}
    \caption{Existing approaches, whether agentic or non-agentic, use language models to obtain context-specific answers and insights, often overlooking the deeper value still embedded in the underlying raw data.}
    \label{fig:agenticmodule}
\end{figure}

\subsection{Contextualization} \label{sec:context}

In enterprise settings, data analytics plays a critical role in generating insights across multiple levels\cite{battle2023exactly}, such as: 1. data observations\cite{1382884,1432690} 2. factual data insights (highlighting patterns like trends, distributions, and outliers)\cite{7106391} 3. hypotheses/question-based insights\cite{7042482}, and 4. domain-knowledge insights, defined as links connecting statistical and/or visual analysis findings with user knowledge\cite{kandogan2018towards,8440860}. Existing research has explored the first three types of insights, leveraging large language models (LLMs) for data analysis and large vision-language models (LVLMs) for visualization interpretation. These technologies have notably enhanced analyst productivity by automating routine analytical tasks. However, little work has been published on domain-related insight generation by LLMs and LVLMs.

Domain-knowledge hits multiple notes. Stemming from psycholinguistics, it not only supplies factual context, but also shapes how information is structured and transformed into insights\cite{mccutchen1986domain}. In enterprise settings, domain-knowledge is defined as the company's operational sphere, shaping the critical questions\cite{evans2004domain}. The assumption here is that domain-knowledge might allow language models to distinguish signals from noise, prioritize metrics synced with business targets, and craft narratives that fit users' mental frameworks\cite{choudhary2024domain, article}.

\subsection{Motivation} \label{sec:motiv}
Current studies on data/business analysis using agentic systems with LLMs or LVLMs mainly zero in on factual data identification and hypothesis/question-driven tasks (QA-pairs)\cite{perez2025llm,wu2024automated,sahu2024insightbench,xu2023chartbench}. Their methodologies emphasize low-level\cite{wu2024chartinsights,wang2025chartinsighter} or high-level analysis\cite{sahu2024insightbench,perez2025llm} aimed at spotting facts or patterns in datasets. This narrow focus leaves a possible gap: 
\begin{quote}
\textit{The limited exploration of how LLMs generate broader insights driven by domain-knowledge through agents.}
\end{quote}
Following this line of investigation, three key questions could be addressed: 1. Will explicit domain identification enhance insights? 2. How do domain-driven insights generated from LLMs differ from those generated by QA tasks? and 3. Can domain-related insights add more meanings to data visualization?

Another challenge we find is how current typical workflows often separate \textit{data-to-chart} \cite{dibia2023lida, wang2023llm4vis, strobel2024hey} and \textit{chart-interpretation} \cite{bendeck2024empirical,choe2024enhancing,wang2024aligned} into distinct areas (For further information, refer to Section \ref{sec:related}). This separation introduces risks, such as the potential disconnection between chart-derived insights and the original data. If charts are inaccurately generated, any insights based on them are inherently flawed, since they rely on erroneous visual interpretations. Additionally, in many business settings, insights may remain hidden within extensive datasets, indicating that merely interpreting chart insights can overlook the broader insights available from the dataset. This drives us to consider the essential need for developing a system that seamlessly integrates the separated workflows.

\subsection{Problem statement}


Building on findings in Sections \ref{sec:context} and \ref{sec:motiv}, the challenge in data analytics with Large Language Model is: 
\begin{quote}
\textit{Developing a robust, generalizable agentic system that uses state-of-the-art LLMs to produce domain-related insights, and enhances data visualization with great significance based on these insights.}
\end{quote}

\subsection{Proposed Solution}
\begin{figure}
    \centering
    \includegraphics[width=1\linewidth]{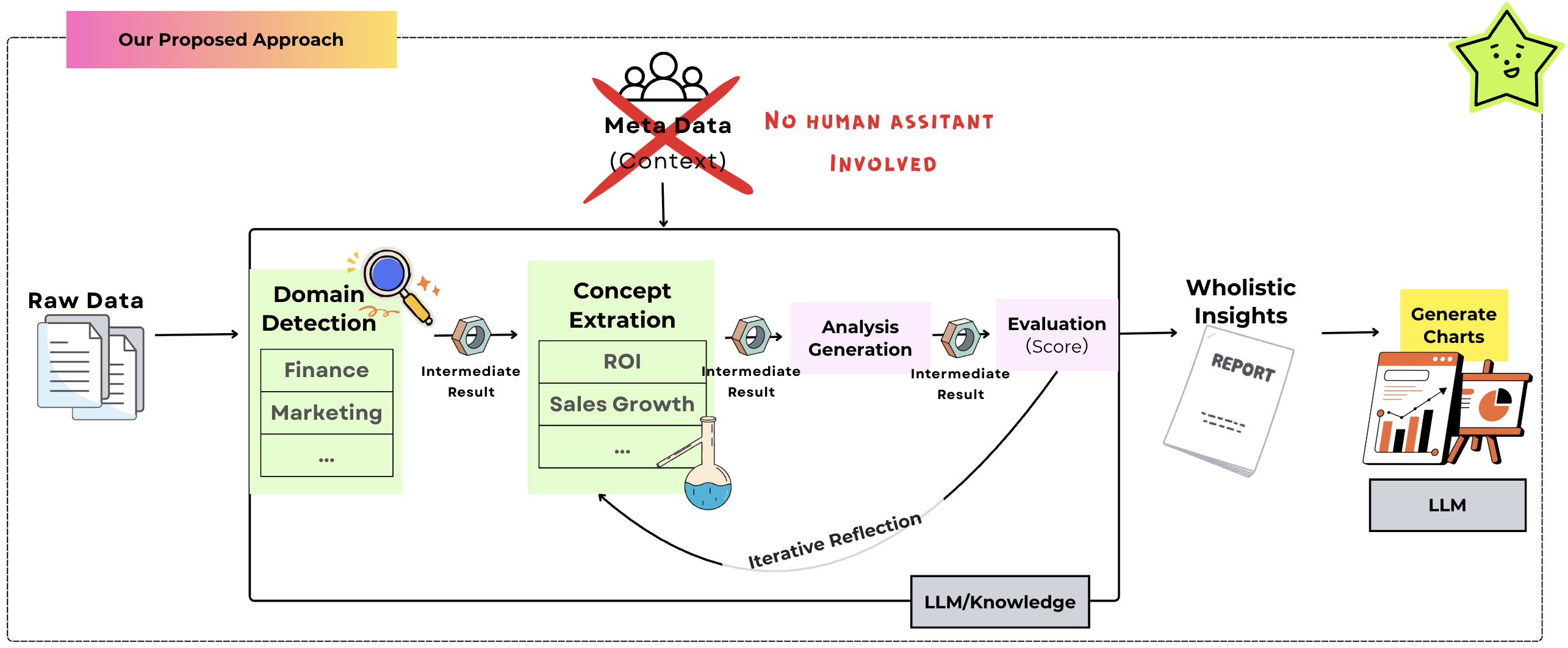}
    \caption{Our end-to-end data-insight-visualization approach provides context-independent domain-aware insights, thus overcoming the limitations of existing systems.}
    \label{fig:ourapproach}
\end{figure}

We propose an end-to-end agentic system that automates the data analysis workflow with a novel pipeline: raw data - domain identification - insights generation - data visualization, a dashboard with charts (See Fig. \ref{fig:ourapproach}, for detailed information, refer to Section \ref{sec:appr}).

Leveraging the capabilities of Large Language Models, our framework consists of specialized agents tasked with: (1) detecting the domain and concepts of the dataset, (2) retrieving and applying domain-relevant knowledge to make insightful analysis, and (3) generating visual insights. Each agent is built with role-specific prompting strategies and is capable of memory-based reflection for iterative improvement. This multi-agent architecture not only grounds the analysis in domain semantics but also enables compositional reasoning, making it adaptable to mixed-domain business datasets. 


\subsection{Contributions}
Below, we list the distinct contributions of our work: 

\begin{itemize}
    \item We introduce a novel end-to-end agent-based framework that transforms raw business tables into insightful dashboards through domain-informed reasoning,
     opening new opportunities for managers in business intelligence, marketing, and finance to validate and refine domain-specific analytics.
    \item We integrate domain detection and knowledge retrieval mechanisms to support feature selection and multi-perspective analytical reasoning, thereby enhancing the accuracy and robustness of the captured insights.
    \item We propose a workflow that simulates the cognitive process of business analysts, enabling iterative improvements and reflective interpretation.


\end{itemize}

\section{Related Work}
\label{sec:related}
Prior work on automated data analysis tasks using LLMs and LVLMs (including data insight generation, chart generation and interpretation) generally falls into two categories: (1) \textbf{data-to-chart} generation, and (2) \textbf{chart-interpretation} summarization. Data-to-chart systems typically target single-table inputs with constrained schema complexity, generating only one or two charts per dataset\cite{wu2024automated}. Such systems tend to emphasize syntactic chart correctness or alignment with pre-specified templates, rather than insight utility. On the other hand, chart-interpretation systems assume an existing chart and aim to summarize it using LLMs\cite{hong2025llms,wang2024aligned}. The resulting insights are usually limited to surface-level features—e.g., maxima, minima, or linear trends—without context-aware reasoning \cite{wang2025chartinsighter}.

Even in more advanced settings, such as domain-specific QA over charts (e.g., in finance), the primary focus remains on factual accuracy or verification against known ground-truths\cite{sahu2024insightbench}. These tasks do not address the cognitive process of deciding \textit{what to visualize} or \textit{what to explore} in the first place. Moreover, very few systems include any notion of domain-knowledge grounding—whether to select relevant metrics, or to contextualize the resulting insights.

Moreover, most existing approaches—whether targeting low-level tasks such as chart factuality check or high-level insight generation—rely heavily on question-answer (QA) pairs\cite{masry2022chartqa,sahu2024insightbench} to drive the analytical process. While this paradigm is effective in settings where users possess prior domain-knowledge and can articulate precise analytical queries, it inherently limits the generative potential of large language models. By anchoring the analysis to predefined questions, these methods constrain LLMs from autonomously exploring the dataset and surfacing insights that may lie outside the bounds of user expectations.

Recent benchmarks such as \textit{InsightBench}\cite{sahu2024insightbench} attempt to move toward higher-level insights. However, their approach remains question-driven in structure with enriched context in metadata as input(See Fig. \ref{fig:agenticmodule}). As a result, the opportunity to leverage the broader reasoning capabilities of LLMs—particularly within agent-based frameworks that simulate open-ended human analytical workflows—remains underexplored.

\section{Proposed Approach} \label{sec:appr}
Our agentic system operates in two sequential stages: data-to-insight and insight-to-chart (Note that the intermediate ``insight'' here refers to domain-knowledge insights. This is in contrast to the question-based insights extracted in in the typical workflows discussed in Section \ref{sec:motiv}). In the data-to-insight stage, the system orchestrates a suite of specialized agents—collaborate to extract semantic and structural understanding from raw tabular data, identifying the business domain, relevant analytical concepts, and candidate insights. The output of this stage provides structured guidance to the next one. In the insight-to-chart stage, the emphasis is not only on syntactic chart generation but also on the production of insightful visualizations tailored to business reasoning tasks.

\subsection{Key Definition}
To support flexible reasoning across diverse business tasks, we introduce two foundational abstractions in our system: \textbf{domain} and \textbf{domain concept}. Rather than imposing a rigid taxonomy, we define these as \textit{relational constructs} that form a hierarchical semantic structure---where a \textbf{domain} denotes a broader business context (e.g., finance, operations, incident management), and a \textbf{domain concept} refers to more granular elements within that context (e.g., revenue, downtime, churn rate). 

This framing intentionally avoids prescriptive definitions or closed ontologies. Inspired by both linguistic theory\cite{mccutchen1986domain} and the generalization capacity of large language models (LLMs), we posit that strict categorization can constrain discovery. Instead, by prompting the LLM to infer these hierarchical relationships on a case-by-case basis, our agentic system encourages open-ended exploration while maintaining enough structure to guide downstream analytical and visualization tasks.

\subsection{Stage 1: Data-to-Insight}
In the \textbf{data-to-insight} stage, the raw dataset serves as the input. Unlike \textit{BenchInsight}, our approach requires zero supplementary context. The output is: a domain name, domain-specific concepts, and analytical insights. This process involves a series of module agents that perform data profiling, detection of domains and concepts, generation of multi-perspective analyses, evaluation, and iterative self-reflection.


\textbf{Data Profiler}~When analysts face large, unfamiliar datasets, a typical approach is to formulate "good questions" that steer a language model toward the desired answers—an idea that underlies much of the question–answer (QA)-driven data-analysis literature \cite{perez2025llm}. Whether the questions probe high-level trends or low-level facts, the workflow is still anchored in asking before seeing. However, our end-to-end agent takes the opposite tack. We begin with an automated data-profiling stage that constructs a structured, statistical synopsis of the table itself, not just a description of data. Using a tree-of-thought prompting method\cite{yao2023treethoughtsdeliberateproblem}, the agent infers column types, value ranges and units, functional dependencies, and potential keys, thereby reducing the ineffective dimensionality of the raw data before any explicit questions are posed. The resulting profile then serves as a principled scaffold for the subsequent reasoning and insight-generation steps


\textbf{Domain Detector}~ This module determines the business theme of the dataset by analyzing its structural profile. Rather than applying rigid classification schemes, it generates a flexible domain label accompanied by a concise, one-sentence definition. This label is inferred using external reference knowledge sources, such as Wikipedia, enabling broad transferability across industries without relying on closed taxonomies or predefined ontologies. The generated domain label is unique to each dataset. In future experiments, as more diverse datasets are tested, inconsistencies or imprecisions may emerge. To address this issue, we propose integrating a self-consistency mechanism—as explored in recent work on multi-round verification using LLMs \cite{liu2025inference}—to enhance stability and balance between generality and accuracy. 
We do not implement this functionality here as this work is a proof-of-concept. 


\textbf{Concept Extractor}~ Given the domain label and structural metadata, the Concept Extractor agent identifies salient concepts that are likely to drive downstream analysis. These concepts are formulated as natural language phrases (e.g., ``monthly active users,'' ``unit cost,'' ``processing delay'') that correspond to the domain and data profiling results. The agent ensures that the extracted concepts are not only relevant to the dataset’s business theme but also actionable for subsequent analytical tasks performed by the Analysis Generator.

\textbf{Analysis Generator}~This agent synthesizes structured insights from the dataset using three lenses: \texttt{descriptive} (summarizing distributions or outliers), \texttt{predictive} (inferring trends or likely outcomes), and \texttt{domain-related}. We pick these lenses because such descriptors offer deeper insight for analysis than non-insight\cite{rony2023augmenting} plus domain-knowledge, highlighted in the motivation, is crucial.

The output is a unified JSON object intended to simulate human-like analysis and hypothesis generation. Emphasis is placed on producing novel, non-obvious insights rather than surface-level observations. 

\textbf{Evaluator}~The Evaluator agent scores the generated outputs across five dimensions: Given that pinpointing the domain and its associated concepts is a crucial cornerstone of our system, the accuracy of domain inference along with the relevance and comprehensiveness of the concepts are key scoring factors. Insightfulness, novelty\cite{lounsbury2021culture} and depth\cite{webb2002depth} are vital to enterprise competence and business analytics. These criteria overlap and agree with recent academic benchmarks and are operationalized into a fixed scoring rubric \cite{perez2025llm}. In addition to numerical scores( ranging between 1 and 4), the Evaluator provides justification for each rating, which gives the next module an enriched textual context to  generate better criticism and reflection.

\textbf{Self-Reflector}~ We adopt the Reflexion framework \cite{shinn2023reflexionlanguageagentsverbal} to enhance reasoning after the evaluation stage. The Self-Reflector receives a composite signal that includes the evaluation scores, contextualized feedback, and memory from prior iterations. The loop runs for up to n iterations or terminates early once all evaluation scores meet a predefined threshold. Notably, we intentionally set a high bar (4 out of 4 across all criteria) to force the LLM to fully utilize its reasoning capabilities, improving insight quality and analytical depth over time. 


\subsection{Stage 2: Insight to Chart}

Our approach implements a Tree-of-Thought (ToT) reasoning framework \cite{yao2023treethoughtsdeliberateproblem} for transforming analytical insights into domain-appropriate visualizations. The ToT methodology was selected specifically to address the complex decision-making inherent in chart element selection, as it enables structured, multi-step reasoning that simulates deliberative expertise. Unlike single-pass approaches that may prematurely commit to visualization choices, ToT facilitates explicit consideration of alternatives and their domain implications. Through the three-expert consensus mechanism, the system can evaluate competing visualization strategies against domain requirements, debate the effectiveness of different chart types, and scrutinize the selection of visual encodings before committing to a final representation.

This deliberative process culminates in a consensus decision specifying the optimal visualization type, rationale, key insight narrative, and recommended annotations that emphasize domain significance. As a result, the system preserves domain-insight context throughout the visualization pipeline, ensuring that the generated charts serve not merely as data representations but as vehicles for communicating substantive domain knowledge.

It is worth noting that the quality of generated insights was found to inversely impact chart generation accuracy. Particularly, when Stage 1 domain insights are effectively transferred to Stage 2, generating accurate charts, especially legends, becomes more challenging.

\section{Evaluation Criteria}

To assess our agentic system's ability to generate insightful visualizations from raw business tables, we adopt a multi-faceted evaluation framework.

\subsection{Textual Insight Evaluation}
\textbf{G-Eval}~We evaluate generated insight outputs using \textit{G-Eval}\cite{liu2023gevalnlgevaluationusing}, adapting their scoring mechanism to better reflect enterprise needs. Specifically, we tailor the prompts to account for \textit{business reasoning insightfulness} and \textit{alignment with domain concepts}, rather than general QA accuracy. To make the evaluation fair, the outputs are scored across (1) \textit{insightfulness}, (2) \textit{novelty}, and (3) \textit{depth}. \cite{perez2025llm}

\textbf{Utilizing InsightBench as a Guide}~ By employing \textit{InsightBench}'s dataset \cite{sahu2024insightbench}, we can assess whether our system successfully determines the proper analytical direction by comparing it with the dataset's ground truth insights. We opted not to adopt \textit{InsightBench} evaluation metric because the workflows of the two systems differ significantly, as illustrated in Fig. \ref{fig:ourapproach}. While the only input our system takes is the raw data, Sahu et. al.'s \cite{sahu2024insightbench} approach handholds the data-to-insights process by incorporating more comprehensive context as input. As such, we can only partially leverage their benchmark in our evaluation.

\textbf{Human Expert Evaluation}~Our framework provides \textit{domain experts and business analysts} the chance to utilize a unified evaluation rubric for human-in-the-loop validation, thereby aligning the evaluation process with practical implementation. In future work, as our work evolves to the next stage, we aim to include professionals from both industry and academia to carry out statistically significant evaluations.

\subsection{Chart Evaluation}
Targeting some Kaggle datasets, numerous data analysts have created and shared their visualization projects. We select a subset of highly downloaded datasets and their corresponding visualizations and analyses, and juxtapose them with our own by: 1. reviewing chart characteristics (type, legend, and axis labels), and 2. evaluating the quality of the insights they contain using G-Eval.

\section{Experiments}
\textbf{Experiment 1: Data-to-Insight}~ This experiment focuses on exploring three essential research questions: 1. Does explicit domain identification matter? This examines the impact of domain labeling on the relevance and depth of the generated visualization (dataset 1\cite{wharton2025customer}). 2. How does our approach compare to a prompt-only baseline? We assess the performance of our agentic pipeline against a baseline that employs the same model, \textit{GPT-4o}, driven solely by a simple prompt, by evaluating differences in analytical insight, novelty, and domain relevance (dataset 1\cite{wharton2025customer}). 3. How does generating insights through questions vary from deriving insights through domain knowledge? We evaluate our approach againsy the latest question-focused study\cite{sahu2024insightbench}, using one of their datasets.

\textbf{Experiment 2: Insight-to-Visualization}~ Using a popular finance survey dataset with download number 19.9K\cite{nitindatta_finance}, we qualitatively examine how our system improves chart generation, analyzing changes in chart type distribution and depth of insights 

\textbf{Dataset Selection}~For broader coverage, we've chosen datasets encompassing diverse domains, data types, and analytical complexities. These include a comprehensive data source from a leading business school's marketing simulation, InsightBench, a classic Kaggle dataset. In future work, a real-world industrial dataset will be examined using our approach.


\section{Results}
\textbf{Result 1: Explicit domain identification matters} 

\begin{figure}
    \centering    \includegraphics[width=0.98\linewidth]{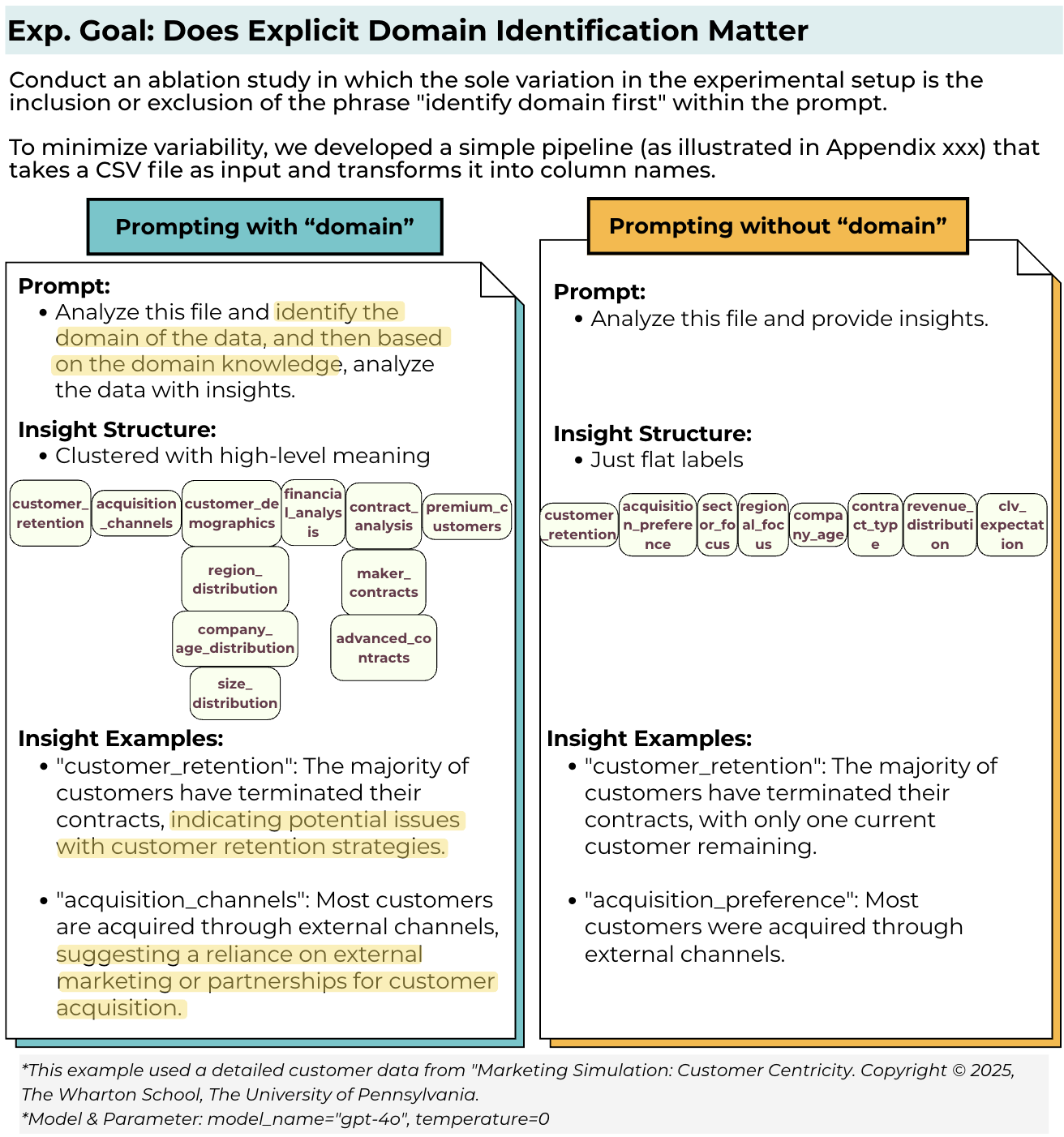}
    \caption{This figure compares the insights obtained with and without domain identification. As can be seen, domain identification grounds the resulting insights, fending it from hallucinations and providing the business analyst with more confidence. }
    \Description{Description of the image for accessibility.}
    \label{fig:domain matters}
\end{figure}

As shown in Fig. \ref{fig:domain matters}, it is evident that adding a simple domain detection instruction to the prompt,substantially improves the generated insights's coverage, structure, and business relevance. 
 Prompting the model to name the domain implicitly triggers a knowledge frame (customer-centric marketing). That frame guides the analysis toward metrics that marketers actually track (LTV, churn cohorts, channel mix), resulting in both deeper topic granularity and hierarchical structuring.
 Furthermore, we notice that while customer churn detection does not hinge on adding domain detection, only the domain-aware run links customer churn to strategy by citing ``potential issues with retention programmes''. 
 Finally, notice that while the baseline run returns broad categories, such as``sectorfocu'' and ``regionalocus'', that are not materially supported by the input raw data, suggesting that the model may be “fishing” for general business axes when no domain context is supplied.
\begin{table}[htbp]
  \caption{Evaluation of our proposed system against non-agentic GPT-4o baseline}
  \label{tab:sys vs 4o}
  \begin{tabular}{cccc}
    \toprule
    G-Eval Metric &GPT-4o & Ours & Relative Lift\\
    \midrule
    Insightful & 0.78 & 0.88 & +12 \% \\
    Novelty & 0.65 & 0.83 & +28 \% \\
    Depth & 0.75 & 0.99 & +31 \% \\
    \bottomrule
  \end{tabular}
\end{table}
This supports our broader hypothesis:
\textit{domain grounding is a lightweight yet powerful prompt-engineering lever for any end-to-end insight-generation pipeline.}

\textbf{Result 2: Our agentic system works better compared to the baseline}. 

As shown in Table \ref{tab:sys vs 4o}, our approach significantly outperforms a non-agentic GPT-4o baseline with domain awareness in terms of insightfulness, novelty, and depth, 

\textbullet{} Insightfulness: Our approach's superior ability to identify nuanced insights, such as the intricate effects of acquisition channels and premium status on customer retention and lifetime value (CLV).

\textbullet{} Novelty: While both our approach and the baseline struggled somewhat in novelty, our system notably surpassed the baseline (0.599 vs. 0.390), successfully generating insights that extend beyond conventional thinking. While these insights may partially correspond to current CRM expertise, since no matter historical or individual novelty are not easy to achieve.

\textbullet{} Depth: Our system outperformed the baseline in terms of the depth of the generated insights (0.942 vs. 0.803), demonstrating that our output exhibits a deeper understanding of CRM complexities, effectively capturing subtle relationships and insights that are implicit in the raw data.

\textbf{Result 3: Insights generated with domain-knowledge can capture the right direction of analysis, 
} 
\begin{figure}
    \centering
    \includegraphics[width=0.75\linewidth]{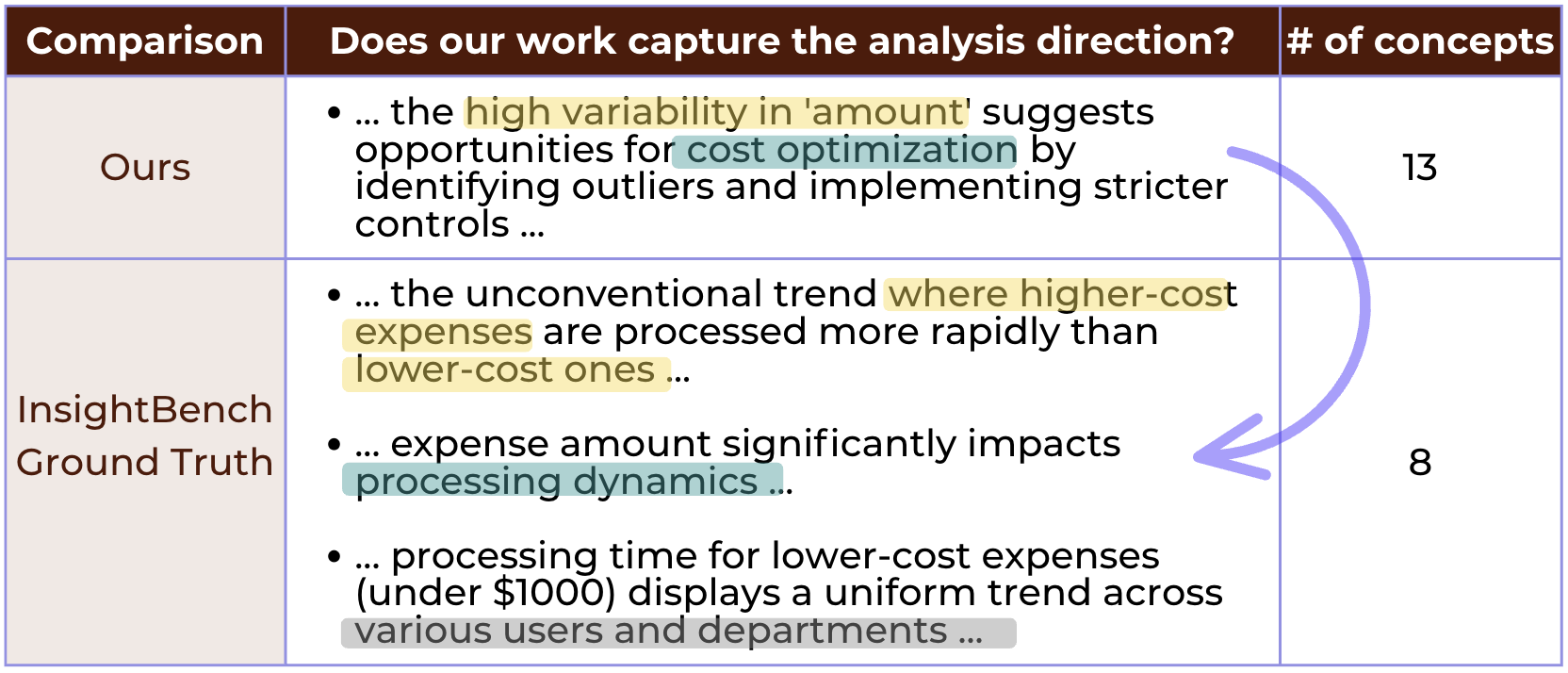}
    \caption{ Comparison of our generated insights with InsightBench ground truth. Our system captures the core analytical direction, identifying key themes such as cost variability, processing dynamics, and optimization opportunities, with broader concept coverage. However, it misses one specific ground truth angle—variation across users and departments—highlighting an area for improvement in capturing organizational structure–related insights}
    \label{fig:insightbench}
\end{figure}

The insights generated with our system successfully aligns with that of \textit{InsightBench}'s ground-truth
, capturing the intended direction of analysis by identifying a key pattern:\textit{"The high variability in 'amount' suggests opportunities for cost optimization by identifying outliers and implementing stricter controls."}.
Specifically, note that our output contains more diverse and forward-looking concepts, especially concerning automation, compliance, and AI integration, while \textit{InsightBench}'s is tightly focused on operational behavior and expense bracket dynamics. 

Moreover, our system highlights several possible analysis directions based on the generated domain-relevant insights, which could be beneficial for broader exploration. As shown in Fig. \ref{fig:insightbench}.
To maintain academic integrity, and due to the need for domain-expert evaluation, we cannot in this work ascertain the other directions' validity. This will be conduct in future revisions.

Overall, our result suggests an important intuition: domain-knowledge is essential for generating an insightful analysis—a hypothesis that could be further tested through causal experiments in future work.

\begin{table}[htbp]
  \caption{A quantitative evaluation of our proposed Data-to-Dashboard method against a Kaggle user\cite{nitindatta_finance_2025}. Notice how our proposed method outperforms the baseline in every single metric.}
  \label{tab:kaggle}
  \begin{tabular}{cccc}
    \toprule
    G-Eval Metric &A Kaggle User & Ours & Relative Lift \\
    \midrule
    Insightful & 0.32 & 0.80 & +147\% \\
    Novelty & 0.34 & 0.61 & +77\%\\
    Depth & 0.36 & 0.77 & +113\%\\
    \bottomrule
  \end{tabular}
\end{table}

\textbf{Result 4: An observation for Human Analyst and Our System} The study's goal is to assess the insightfulness of chart creation by juxtaposing our analysis with that of the Kaggle user who received the most "Upvotes" on a personal finance survey dataset\cite{nitindatta_finance}. First, we employed G-Eval to assess the insights described in the Kaggle user's work \cite{nitindatta_finance_2025}, as well as the caption in its generated chart
. To evaluate the results in their proper context, we modified the evaluation prompt's criterion from "deep domain expertise" to “Does the analysis demonstrate deep insights?” As shown in Table \ref{tab:kaggle}, our system significant outperform on this analysis task. 

\begin{figure}
    \centering
    \includegraphics[width=1\linewidth]{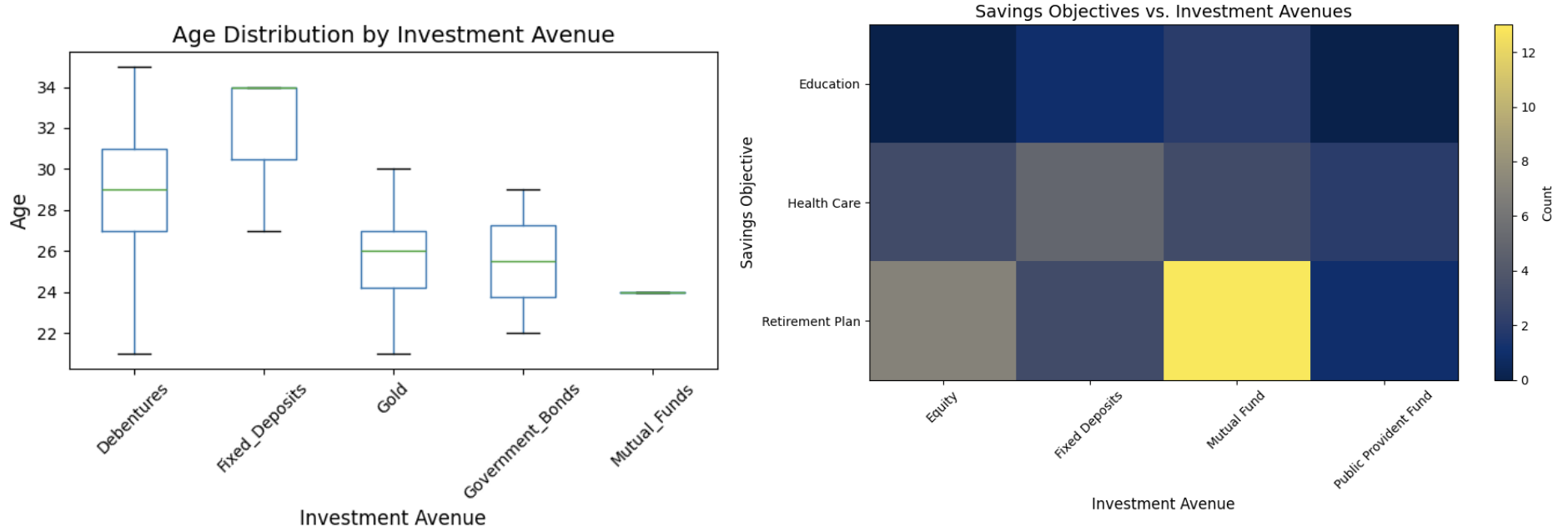}
    \caption{Examples of insightful figures generated by our approach}
    \label{fig:charst}
\end{figure}

Secondly, we identify the chart types produced by Kaggle users, which includes 8 bar charts and 1 box plot. In our own work, we created 1 stacked bar chart, 1 heatmap, 1 pie chart, 1 scatter plot, and 1 box plot. Despite the incomplete development of our stage 2 insight-to-chart agentic system due to time constraints, resulting in 2 charts being plotted incorrectly due to coding errors, our work still demonstrates highly insightful chart attributes, shown in Fig. \ref{fig:charst}.

\section{Acknowledgments}

We would like to thank Prof. Samuel Engel at Boston University, for generously providing several valuable datasets that supported this project. We would also like to thank Yuan Gao, for the early-stage discussions and paper recommendations that helped shape the initial direction of this project.

\bibliographystyle{ACM-Reference-Format}
\bibliography{sample-base}

\end{document}